# Experimental Analysis of Trajectory Control Using Computer Vision and Artificial Intelligence for Autonomous Vehicles

Ammar N. Abbas
*Mechatronics Engieering Department*
*School of Engineering*
Nile University, Giza, Egypt
ammar.abbas@eu4m.eu

Muhammad Asad Irshad
*Mechatronics Engieering Department*
*School of Engineering*
Nile University, Giza, Egypt
asad.irshad@eu4m.eu

Hossam Hassan Ammar
*Smart Engineering Systems Research Center*
*School of Engineering and Applied Sciences*
Nile University, Giza, Egypt
hhassan@nu.edu.eg

***Abstract*—Perception of the lane boundaries is crucial for the tasks related to autonomous trajectory control. In this paper, several methodologies for lane detection are discussed with experimental illustration: Hough transformation, Blob analysis and Birds eye view. Following the abstraction of lane marks from the boundary, the next approach is applying a control law based on the perception to control steering and speed control. In the following, a comparative analysis is made between an open loop response, PID control and a neural network control law through graphical statistics. To get the perception of the surrounding a wireless streaming camera connected to Raspberry Pi is used. After pre-processing the signal received by the camera the output is sent back to the Raspberry Pi that processes the input and communicates the control to the motors through Arduino via serial communication.**

## I. Introduction

Self-driving cars can be classified in four technological sub-classes as per function and automation equipment used to replace human driver. That is environment perception, navigation, path planning, and control system [1]. Navigation system's core task is to determine current location of car and assess path trajectory to reach destination. Matching the human intelligence for navigation and planning, self-driving cars must be intelligent enough to solve this problem autonomously. Different techniques can be utilized to tackle this issue of navigation, ex. Car navigation system based on GPS and GIS, location system determine on the bases of relative, absolute and hybrid location. Electronic Map is digital map that contains information regarding geographically, traffic concentration based, building and road facilities. Environmental perception is 2nd most important feature of self-driving cars AKA Mobile robots. Mobile robots (MR) traverse trough static or dynamic environments. Radar perception, laser or lidar perception [2] and visual perception such as event-based; bio inspired cameras [3] serves the navigation task with unique control decision capabilities. This feature enables self-driving cars to generate perception of surrounding environment such as lane detection, sign recognition, obstacle avoidance and enable it to perceive independently. In static environment MR is require traversing from starting point to end point on permissible path, while optimizing the distance, time and energy consumed as well as avoiding obstacles. While dynamic environment requires better perception and intelligence to make decision process robust for changing environmental scenarios. The main task remains same as of static environment to optimize distance, time, and energy consumption and avoid obstacle while analyzing abrupt changes in surrounding environment. An adaptive control strategy will be implemented in this research work to solve optimization problem.

## II. Literature Review

Doke Akshay M et al [4] have concentrated on the use of autonomous vehicle, in which the vehicle will identify an obstacle, traffic sign and stop signs. It likewise drives itself on the track for checking and reconnaissance the environment with the assistance of Ultrasonic sensors, IR sensors. Arduino, Raspberry Pi which is utilized to gather all the information from sensors and cameras. Claes, Daniel et al [5] proposed the development of a robotic car with hardware control, lane detection, mapping, localization and path planning capabilities. Aim of this work was a completely independent, reliable and robust system that can traverse a single lane track bordered by white lines on an optimal path. To detect the track boundaries, two different approaches were implemented. A RANSAC approach, which approximates the lines by random sampling of splines, and a Polyline approach, which applies primitive image processing in combination with a road model. To map the environment, odometry and vision-based information was fused by a particle iteration based Simultaneous Localization and Mapping system. The map was afterwards used in conjunction with Adaptive Monte Carlo Localization. For path planning, a one-step continuous-curvature approach based on sensor or maps data was used. Aditya Kumar et al [6] proposed a working model of self-driving vehicle which was equipped for driving from one area to the next or to remain on various sorts of tracks, for example, curved tracks, straight tracks and straight pursued by curved tracks. A mounted camera module over the highest point of the vehicle alongside Raspberry Pi sent the pictures from real world to the Convolutional Neural Network which at that point anticipated one of the accompanying directions. Right, left, forward or stop. Which was then trailed by sending a signal from the Arduino to the controller of the in the desired direction without any human intervention. Truong-Dong et al [7], proposed a monocular vision-based self-driving vehicle model utilizing Deep Neural Network on Raspberry Pi. In this work, focus is on finding a model that directly maps raw input images to a predicted steering angle as output utilizing using a deep neural network. To begin with, the CNN model parameters were trained by utilizing information gathered from vehicle platform built with a RC car, Raspberry Pi computer and front facing camera. The training data were road images combined with the time-synchronized steering angle generated by manually driving. The test results show the adequacy and power of autopilot model in lane keeping task. Haji et al [8] have utilized the idea of capturing the cognitive skills of human and regenerating it in a computer algorithm in order to transform a normal scaled version of RC car model to an autonomous car with the help of deep learning methodology. Dhariwal et al [9] replicated a real world self driving car scenario with the use of an RC car using neural networks. The proposed methodology uses the sensor information from the camera and an ultrasonic sensor that sends data through raspberry pi to the computer wirelessly, rest of the computation is done on the computer including the decision making process by the neural network and the processed decision is then sent to arduino which is used to control the acceleration and the direction of the model car. Maciek Dziubiński [10] built a similar project, an RC model of the car trained on the real world scenario using the deep neural networks with the heuristic based on the depth camera installed in the front of the car. The processing was done on NVIDIA Jetson GPU. the Jetson module then communicated with another microcontroller Teensy-LC which is used to control the steerig of the car based on the decision made by the neural network. WeiHang Feng et al [11] uses the concept of autonomous driving in education purpose. The paper proposes a simulation for the model along with the scaled version of the car for testing the algorithm that is trained using the deep neural networks. The model is trained using 813,025 total number of parameters based on Keras/ TensorFlow. The parallel testing interface is developed for the simulated and the real world environment. In some approaches [12, 13], authors uses deep reinforcement learning approach to teach obstacle avoidance tasks to the car without manual training provided by the labeled data.

## III. Experimental Setup

The hardware and electronic schematic is described in figure 1. Wireless Simulink communication was used for real-time hardware in loop control as well as to deploy the code to Raspberry Pi (Rpi). Two motors are used to control the system, for linear motion Brushless DC (BLDC) motor and for steering a servo motor is utilized. Rpi performs the computational process because of its high computational efficiency and Arduino is used for the hardware control.

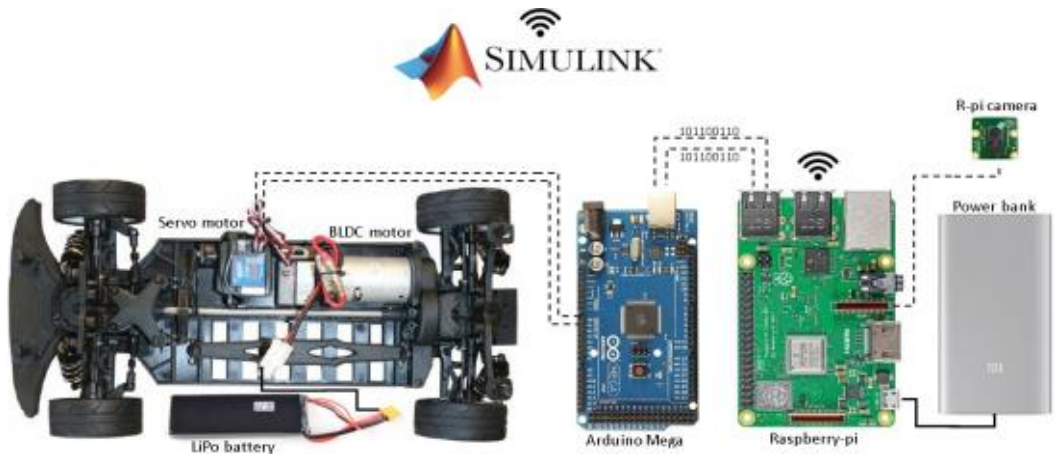

Fig. 1. Hardware and electronics schematic

### A. Testing Environment

A noise-free testing environment was used for the experimental analysis. The path to be followed autonomously by the car was first constructed on Matlab app to run and simulate the response using ”Driving scenario designer”. The pre-programmed points were defined along the path from which a rear camera view and a bird-eye plot is generated as shown in figure 2.The front sensor’s active area is also plotted on the real-time through which the lane detection algorithm generates line plots between world and system coordinates. The project lab provided by the Nile university was used to perform the experiments where a basic lane model was developed on the ground using black electric tape as shown in figure 3.

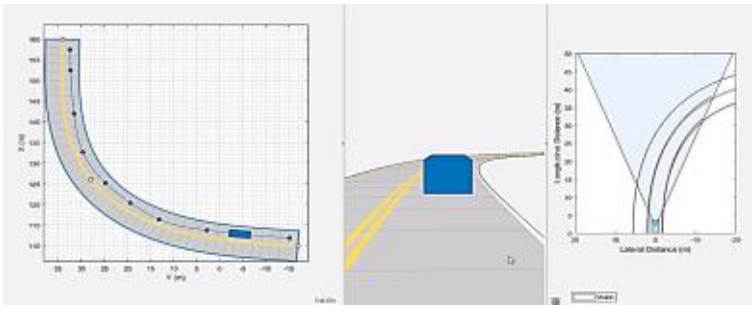

Fig. 2. Simulation model using ”Driving Scenario Designer”, MATLAB

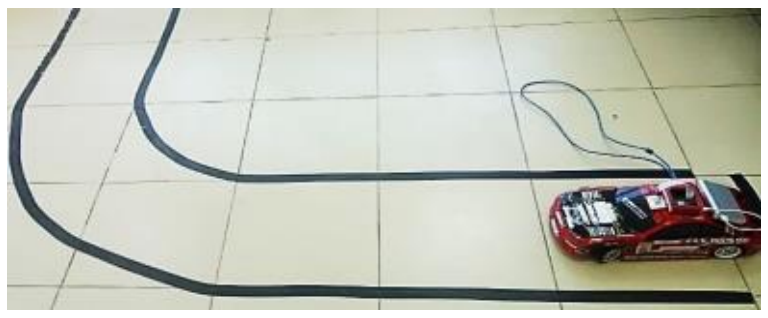

Fig. 3. Lane marker using black electric tape on ceramic tiles

## IV. Methodology

The user-interface for programming and visualizing is connected to the vehicle-interface through a wireless communication within Simulink. The sensing elements for location is provided by the Rpi camera along which to control the velocity an electronic speed controller is wired between the source and the motor. To track orientation and heading, steering servo is used for the feedback. After receiving the necessary elements to sense the surrounding, the next phase is to perceive autonomously. With the visual sensor, lane boundaries gives the perception of path to follow. After perceiving and decision making, the planning and control parameters are set by defining global and local control. Global control determines the heading and velocity for the determined path, however, local control is updated over every instant to take over the global control in case of varying environment such as obstacles, pedestrians or change of path etc. The final control parameters are sent by the controller to the vehicle in form of steering angle and controlled velocity.

### A. *Flowchart*

The flow diagram of this experiment is defined in figure 4. In order to set-up hardware-in-loop configuration, a wireless communication between the programming platform or user is initiated with the controller. After initializing the system, live video is sent from the camera to the controller for processing. The processing begins with defining a Region of Interest (ROI) within the surrounding that contains the useful information. For the specified ROI, the image is processed in order to extract the lane boundary information and converting it into the vehicle coordinates. If all the check fails then the final point of obstacle detection or end of lane boundary (dead-end) have been detected and the loop stops. The following subsections will describe in detail the flow sequentially.

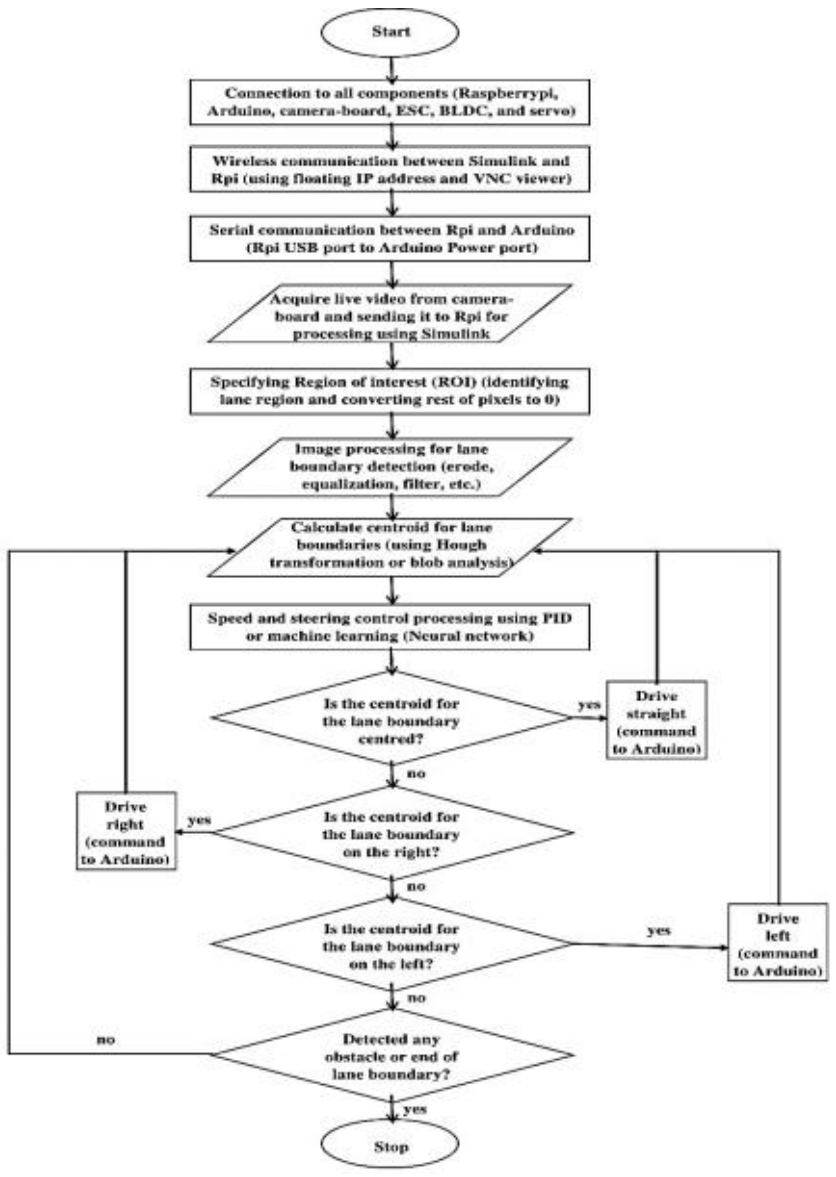

Fig. 4. Flow diagram

### B. *Simulink Block Diagrams*

The block diagram using Siumulink was created for the autonomous control of the car that was deployed on Rpi as shown in figure 5. Steering and speed control open loop response was performed through deflection calculation from one of the two input methods that will be discussed in the following sections.The data was sent and received to and from Arduino through serial communication [14].

The subsystem for the system described in the preceding section is expanded in figure 6. Image is acquired from Rpi camera with a series of continuous images forming a real-time video. The acquired data is sent for further processing and the final output is fed for extracting lane information and calculating deflection from the desired path. The analysis is then sent to the controller for lane tracking.

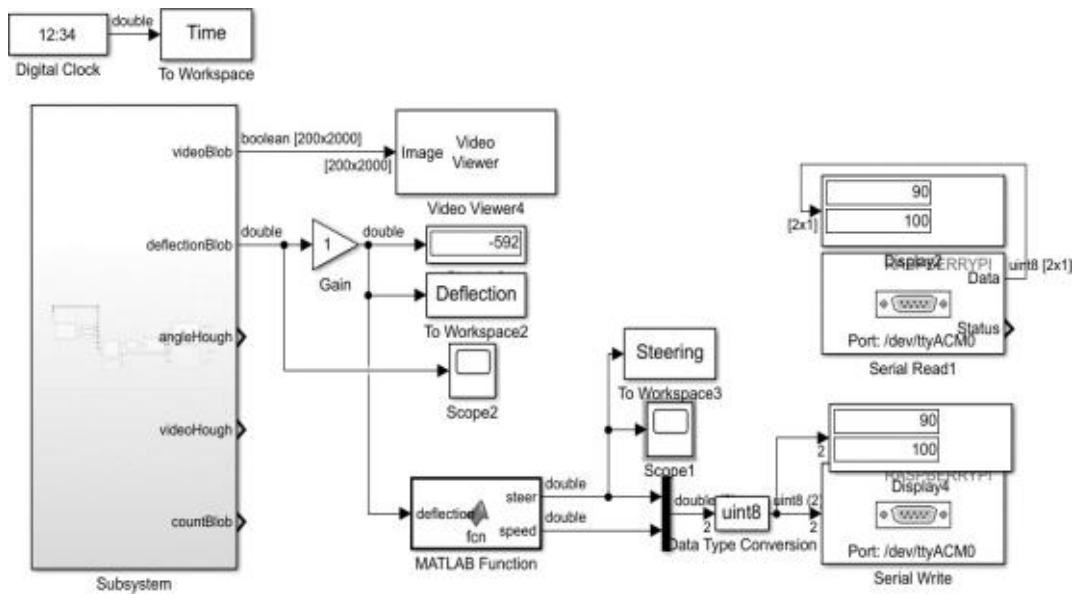

Fig. 5. Simulink block algorithm to control the autonomous car

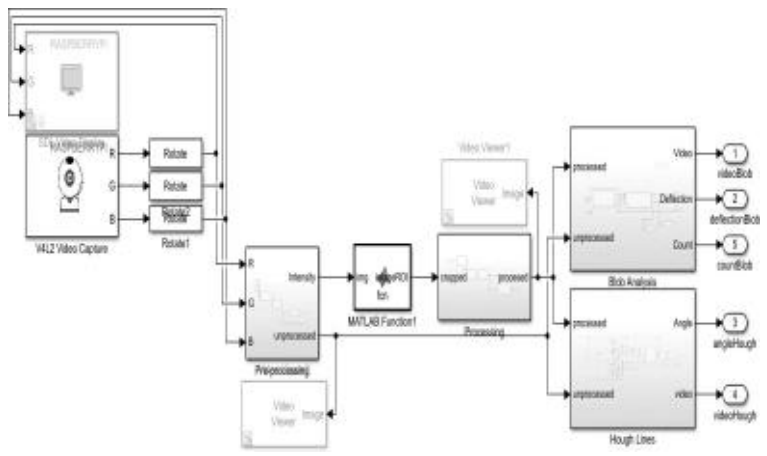

Fig. 6. Subsystem for acquiring and processing image

*1) Pre-processing:* Before further processing, it was required to convert the RGB image into an intensity image that is passed through primary adjustments to equalize the pixel values and contrast in order to elucidate the important information as shown in figure 7.

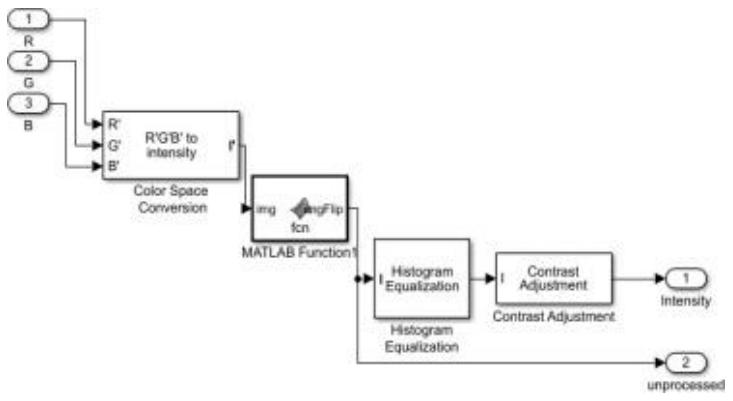

Fig. 7. Pre-processing block

*2) Processing:* The region of interest (ROI) of the pre-processed image is created in order to filter and process just the necessary information. To eradicate the noise, erode function with different parameters are used twice with targeting specific noise. The median filter removes the remaining noise. The filtered image is passed through auto-threshold to binarize the image into black and white image that explicate the black lanes in the testing environment as shown in figure 8.

The acquired and processed image are shown in figure 9. The image is pre-processed to equalize the pixel values and normalize any noise related to bright lights and reflections. To detect just the lanes, the background pixels were converted to zero and a specific region was located to perform the processing as shown. The important information of the lane is remained that shows if the lane is straight or curved and can estimate the deflection angle from the desired path.

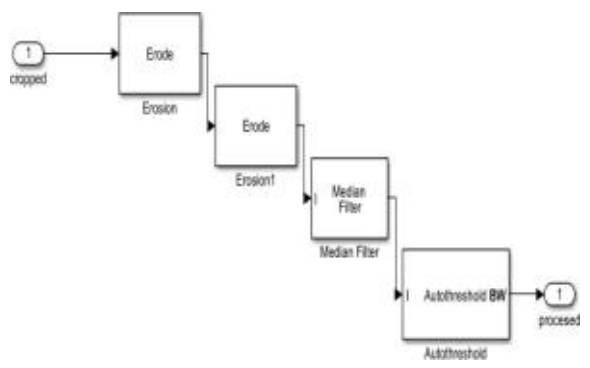

Fig. 8. Filtration

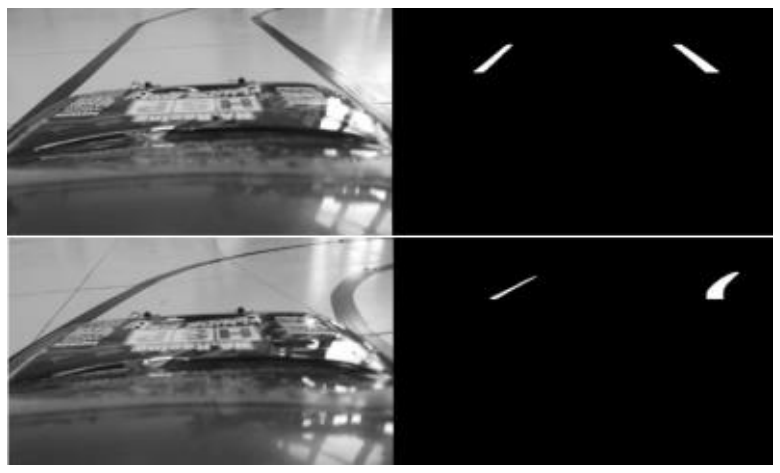

Fig. 9. Unprocessed and processed grey-scale image from camera

*3) Blob analysis:* The processed image with the specified ROI is ready for extracting information about lane boundary points. Here two methods are proposed: first is blob analysis. The Simulink block is used to automatize the process and create centroid, boundary box and orientation of the blob as shown in figure 10. A Matlab function is then created which calculates the end points of the blob to calculate the average and draw an imaginary line in between the two lanes. The slope of that line determines the magnitude of deflection from the desired path with respect to the actual path.

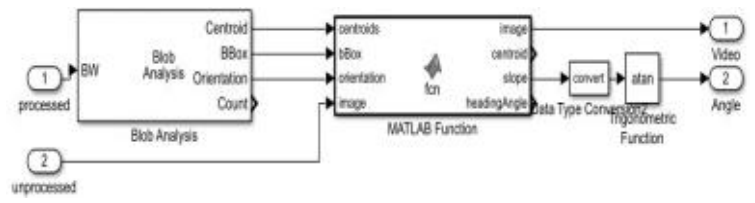

Fig. 10. subsystem for blob analysis

The processed image passed through blob analysis is shown in figure 11. The average points are located from the extreme corners of the box around the blob. The two points serves the function of calculating slope which is then converted to an angle. The magnitude of left and right turns are detected by those angles in positive and negative directions which is then passed to the controller to determine the required steering angle and speed. The output for motors is then sent to Arduino.

*4) Hough lines:* The second approach was by using Hough lines transformation to calculate local maximas and generate line points over the lane boundaries as shown in figure 12. The generated lines through continuous points was then passed to a mean block creating an average line in between the boundaries. The slope of the averaged line was calculated to determine the deflection scale.

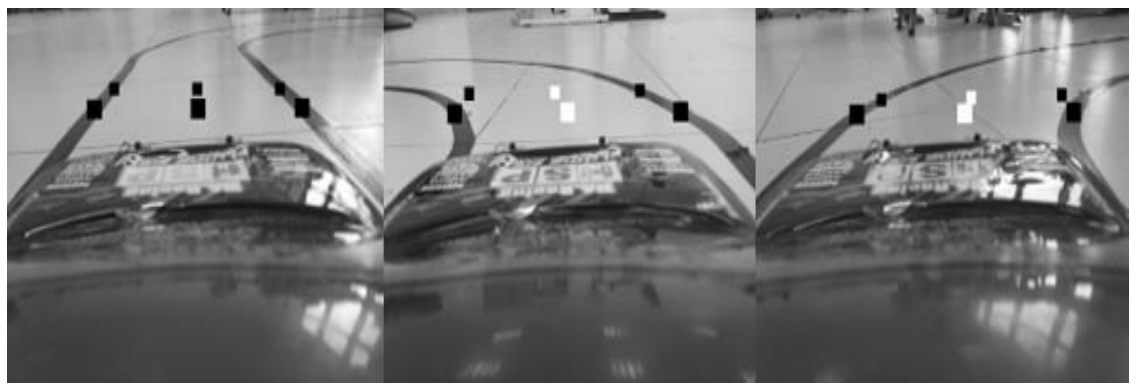

Fig. 11. (i) Blob analysis: centroid detection and bounding box markings

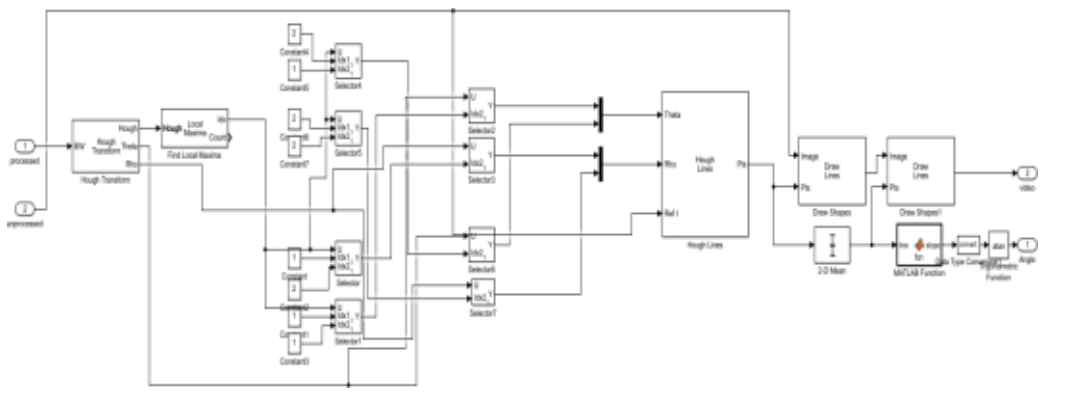

Fig. 12. subsystem for hough transformation

The processed image passed through hough transformation is shown in figure 13. The average line is generated from the two lane boundaries. The angle of the mean line is determined which determines the required steering angle in the same way as described in the previous part.

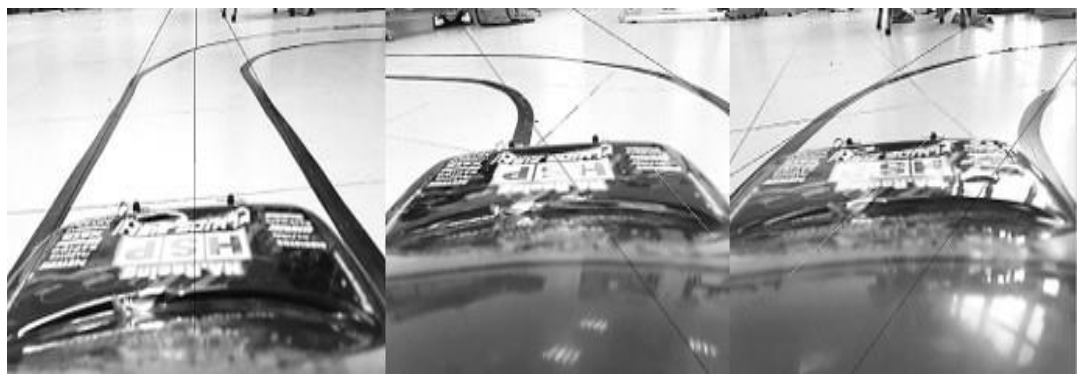

Fig. 13. (ii) Hough lines: line detections through local maxima calculations

*5) Bird's eye view:* Another method was also used to track the lane boundary through Matlab algorithm called Bird's eye view [15]. The first step for which was to extract out the camera parameters in order to get the lane points in camera's frame of reference through the world's frame of reference. The camera calibration app was used to obtain the intrinsic and extrinsic parameters of the camera with the aid of checker-box image. Several images were obtained by placing the board in front of the camera and changing the location and orientation manually while the images were being stored in a folder. The images were then used in the app to calculate the checkers and the calculation error through which the parameters were calculated. Fig 14 shows the calibration and images from camera's frame of reference.

Finally, a test algorithm was used to test the lane boundary detection by placing checker-box image in the test area in order to get the conversion parameters. A birds eye view was then generated using the tuned parameters which was then further filtered to detect lane. The boundary points generated from world coordinates were converted to the camera and car coordinate system which was then used to create a line over the points as shown in figure 15.

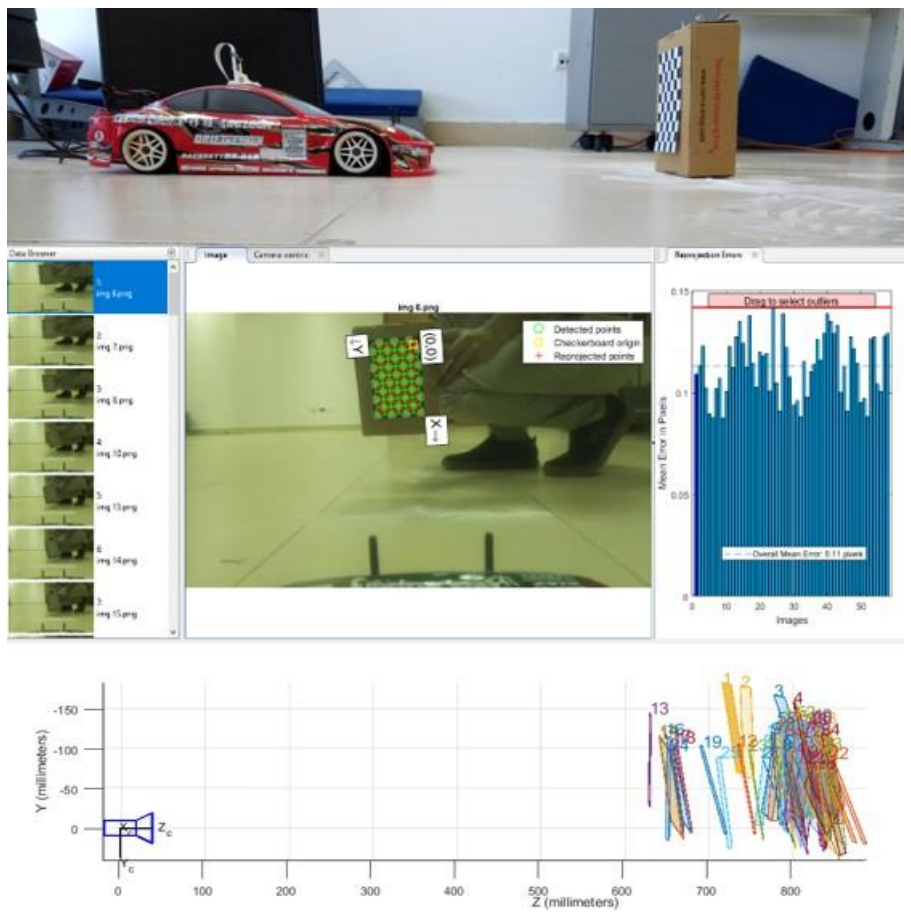

Fig. 14. Camera calibration

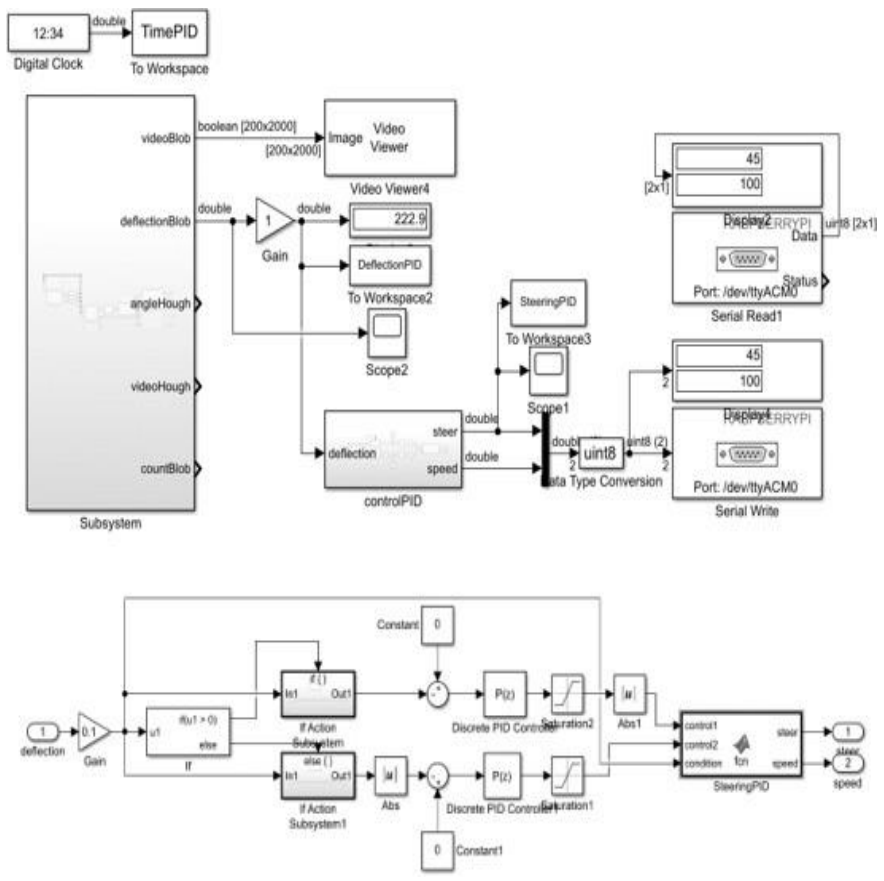

Fig. 16. (PID control law

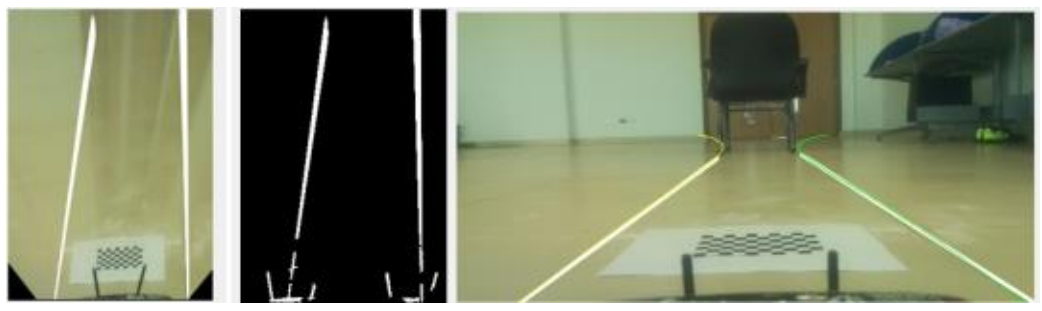

Fig. 15. Lane boundary detection through Bird's eye view

## C. Control laws

Out of the three methodologies proposed in order to detect lane boundaries just "blob analysis" was used for comparison between the control laws. The first step was to test the results through a simple if else statement loop without using a control law. After getting results from the open loop control system, two control strategies were adopted. (i) PID and (ii) Neural networks. Both controls are tested and compared. In order to get data points to plot graph as well as to train the neural network a digital clock block was used to send continuous time data to work space and the output for deflection and steering were stored in arrays. The control blocks were inserted between the deflection measurement (input) and steering plus speed values (output) that were sent serially to arduino through Rpi.

1) *PID:* The PID control subsystem is shown in figure 16. The two PID loops are used to separate the positive and negative parts of the control loop. A saturation threshold is added to fix the minimum and maximum intervals depending on the servo signals required for the motors. The positive and absolute of the negative value at the end is then fed into the Matlab function block that on the basis of the output from PID condition generates an output for controlling the motors.

2) *Neural network:* After getting the results from PID, next step was to train the system through neural networks and apply the trained network block between the input to generate the desired output. The input data was generated by manually controlling the car through an android application communicating through a Bluetooth module. The input for deflection along the lane boundary was generated through the camera and the output was trained by the real time control of the car through the manually controlled values of steering and speed. The trained block schematic used in the network is shown in figure 17.

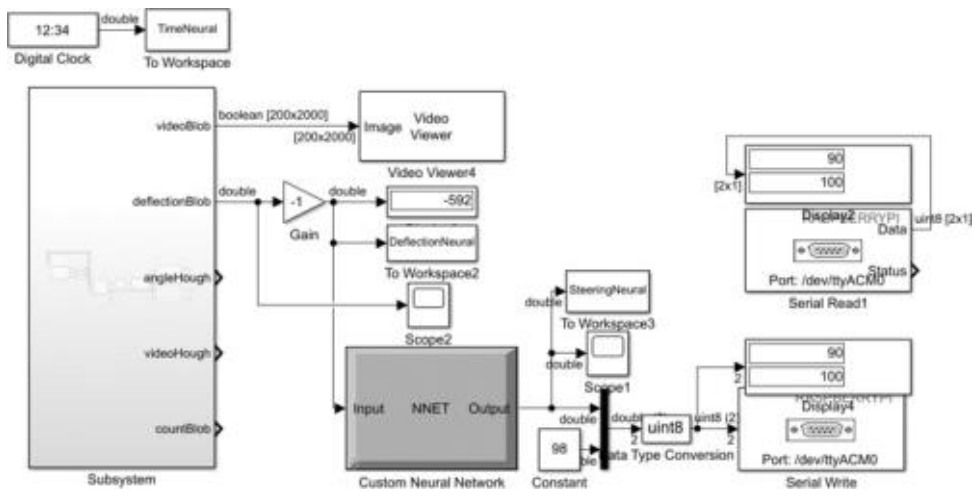

Fig. 17. Neural network control law

## V. Results

The data was stored in an array and the stored data is shown graphically for each test (open loop, PID and neural) in figure 18. The input (deflection) and output (steering) of the graphical representation is further used to quantify the comparison through statistical representation on graph.

The statistical data from each graphical representation is shown in table 1. For which the steering data for open loop response (normal) is just 0 and 90 for left and right motion respectively. however, for other procedures the value is in between 0 to 90 depending on the deflection (input). By comparing the standard deviation for three controls it can be seen that the minimum deviation exists in neural for both the deflection as well as the steering control.

Figure 19 shows image shots from the video made during the test. The drive is completely autonomous and the steering control can be seen at the curve. and figure 20 shows the real time camera shots through an on-board Rpi camera sending data to the computer through Simulink.

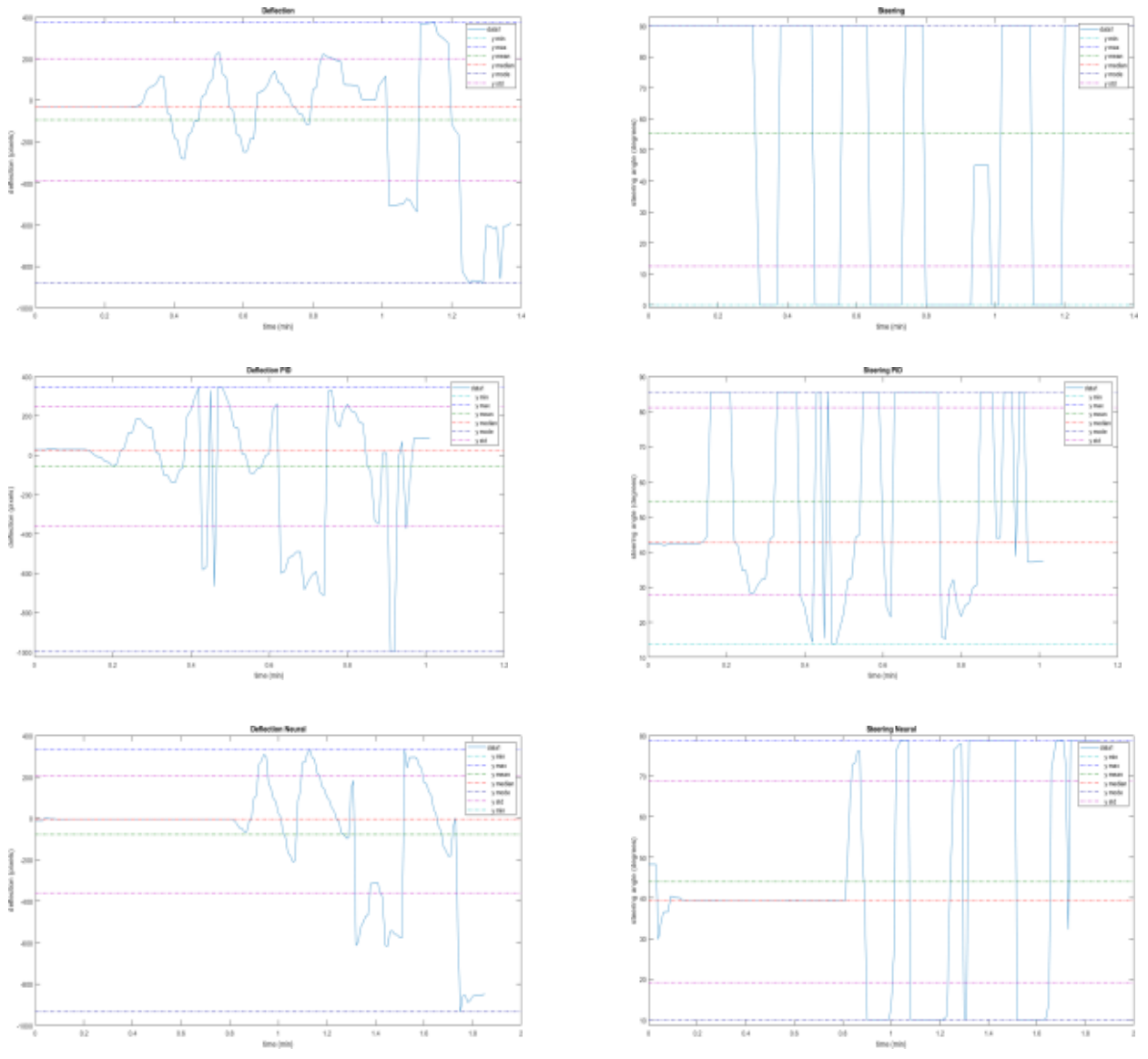

Fig. 18. Steering and deflection graphs for open and closed loop responses

| Statistics | | | | | | |
|---|---|---|---|---|---|---|
| Params | Deflection | | | Steering | | |
| | Normal | PID | Neural | Normal | PID | Neural |
| Min | -880.48 | -1000 | -933.23 | 0 | 13.75 | 10.00 |
| Max | 376.08 | 347.17 | 332.08 | 90 | 85.50 | 78.86 |
| Mean | -95.20 | -55.80 | -77.93 | 55.43 | 54.47 | 43.98 |
| Median | -33.13 | 26.07 | -3.86 | 90 | 42.65 | 39.40 |
| Mode | -880.48 | -1000 | -933.23 | 90 | 85.50 | 10.00 |
| Std. dev. | 294.73 | 303.65 | 282.15 | 42.91 | 26.68 | 24.80 |
| Range | 1256.6 | 1347.2 | 1265.3 | 90 | 71.74 | 68.86 |

TABLE I
COMPARISON BETWEEN OPEN AND CLOSED-LOOP RESPONSES

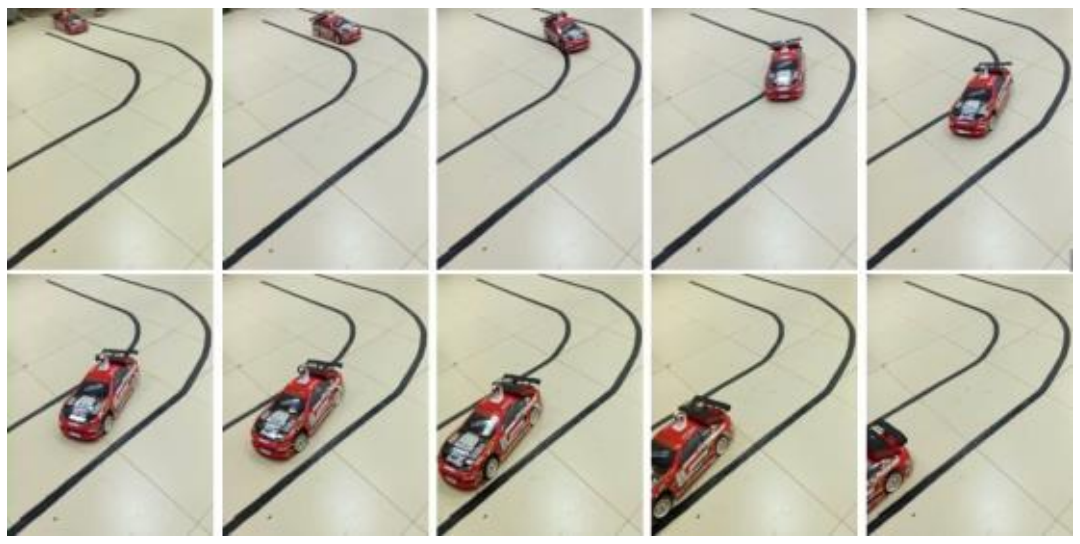

Fig. 19. Real time image shots while testing

## VI. CONCLUSION AND FUTURE WORK

The open loop response and closed loop response with PID and neural network defined as the control laws were discussed in this paper for autonomous lane following cars. Deflection from the center of the lane and the related steering command were stored and displayed graphically. All the responses were statistically compared for quantifying the results. The results with neural network were more promising based on the standard deviation for each graphical results. All the tests were based on the methodology of detecting the lane through blob analysis. However, future work will be associated by testing the car through detecting lanes by other methods as well, such as, Hough analysis or Bird's eye view which were also discussed in the paper's methodology section. Furthermore, the aim is to compare the results of these three methodologies for each control law.

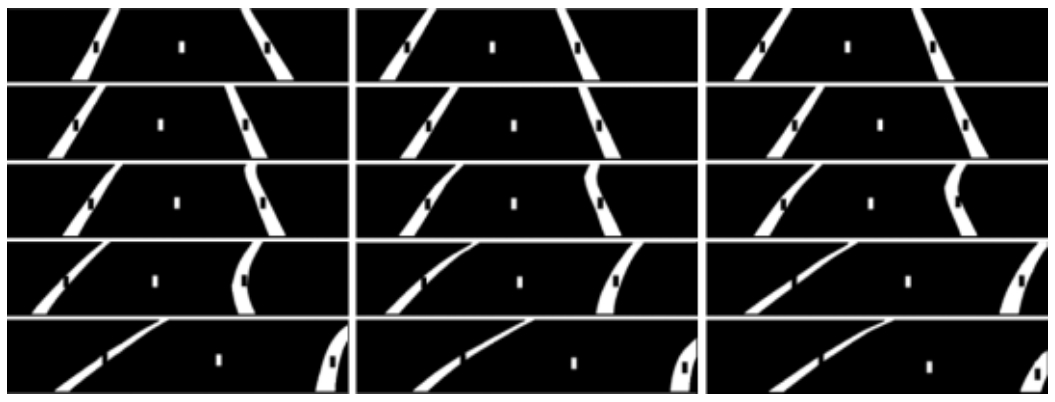

Fig. 20. Camera view real-time shots while testing